\documentclass[10pt,twocolumn,letterpaper]{article}

\usepackage[pagenumbers]{wacv} 
\usepackage{array}
\usepackage{graphicx}
\usepackage{float}
\usepackage{multirow}
\usepackage{makecell}

\definecolor{wacvblue}{rgb}{0.21,0.49,0.74}
\usepackage[pagebackref,breaklinks,colorlinks,allcolors=wacvblue]{hyperref}

\def\wacvPaperID{2000} 
\def\confName{WACV}
\def\confYear{2026}

\title{Segmentation-Aware Latent Diffusion for Satellite Image Super-Resolution: Enabling Smallholder Farm Boundary Delineation}

\author{Aditi Agarwal$^{1}$
\and
Anjali Jain$^{2}$
\and
Nikita Saxena$^{1}$
\and
Ishan Deshpande$^{1}$
\and
Michal Kazmierski$^{1}$
\and
Abigail Annkah$^{3}$
\and
Nadav Sherman$^{3}$
\and 
Karthikeyan Shanmugam$^{1}$
\and
Alok Talekar$^{1}$
\and
Vaibhav Rajan$^{1}$\\
$^{1}$Google DeepMind, $^{2}$Google, $^{3}$Google Research\\
\normalsize \texttt{aditie@google.com}
}

\newcommand{\cmark}{\textcolor{green!60!black}{\checkmark}}
\newcommand{\xmark}{\textcolor{red!80!black}{x}}

\begin{document}
\maketitle

\begin{figure*}[htbp]
\captionsetup[subfigure]{labelformat=empty}
\captionsetup{font=footnotesize}
\centering
    \begin{subfigure}[b]{0.18\textwidth}            
            \includegraphics[width=\textwidth]{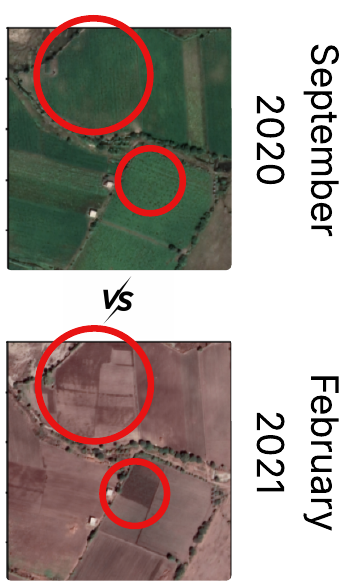}
            \caption{}
            \label{fig:SRl}
    \end{subfigure}%
    \begin{subfigure}[b]{0.73\textwidth}
            \centering
            \includegraphics[width=\textwidth]{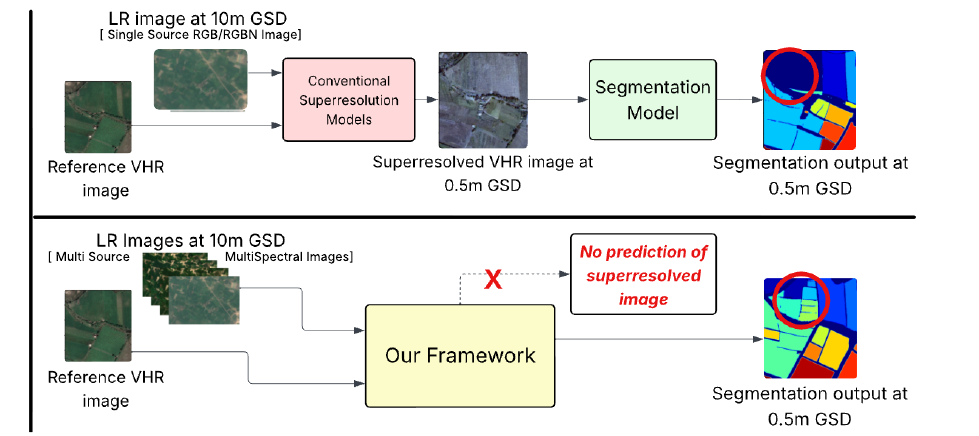}
            \caption{}
            \label{fig:D-Imager}
    \end{subfigure}
    \vspace{-0.5cm}
    \caption{ (Left) Example of subtle temporal changes in agricultural field boundaries. (Right) SEED-SR (below) directly generates high-resolution segmentation from an older VHR reference and a recent LR temporal stack, unlike current (above) 2-step super-resolution and segmentation. Best viewed in colour.}\label{fig:TOF}
    \vspace{-0.4cm}
\end{figure*}
\begin{abstract}
Delineating farm boundaries through segmentation of satellite images is a fundamental step in many agricultural applications.
The task is particularly challenging for smallholder farms, where 
accurate delineation requires the use of high resolution (HR) imagery which are available only at low revisit frequencies (e.g., annually). 
To support more frequent (sub-) seasonal monitoring, HR images could be combined as references (ref) with low resolution (LR) images -- having higher revisit frequency (e.g., weekly) -- using reference-based super-resolution (Ref-SR) methods.
However, current Ref-SR methods optimize perceptual quality and smooth over crucial features needed for downstream tasks, 
and are unable to meet the large scale-factor requirements for this task. 
Further, previous two-step approaches of SR followed by segmentation do not effectively utilize diverse satellite sources as inputs. 
We address these problems through a new approach, 
\textbf{SEED-SR},
which uses a combination of 
conditional latent diffusion models and 
large-scale multi-spectral, multi-source geo-spatial foundation models.
Our key innovation is to bypass the explicit SR task in the pixel space and instead perform SR in a segmentation-aware latent space. 
This unique approach enables us to generate segmentation maps at an unprecedented 20$\times$ scale factor, and
rigorous experiments on two large, real datasets demonstrate up to \textbf{25.5}\% and \textbf{12.9}\% relative improvement in instance and semantic segmentation metrics respectively over approaches based on state-of-the-art Ref-SR methods.
\end{abstract}
    
\section{Introduction}
\label{sec:intro}

Satellite remote sensing provides invaluable data at a global scale in many scientific domains.
In agriculture and environmental studies, it can drive improvements in
multiple applications such as 
land cover mapping \citep{dua2024agriculturallandscapeunderstandingcountryscale},
crop identification \citep{agilandeeswari2022crop}
agricultural monitoring \citep{murugan2017development} and crop yield estimation \citep{roznik2022improving} and
agricultural drought monitoring \citep{Skakun2015}.
Effective solutions to these problems can have significant impact on global food security and environment \cite{inbook}.
A foundational step in several agricultural applications is \textit{field boundary delineation}, the task of predicting the polygonal boundaries and constituent areas of crop fields from overhead satellite images \citep{WaldnerDiakogiannis2020Deep,MeiEtAl2022Using,KernerEtAl2023Multi}.
The problem is especially challenging for \textit{smallholder farms}, containing fields smaller than 2 hectares \citep{fao2005economic}; and is crucial for the Global South comprising about two-thirds of the developing world’s 3 billion rural people in about 475 million 
households \citep{inbook}.

The task is formulated as an image segmentation problem and several deep learning models, developed for natural images, have been adopted for satellite images \cite{WaldnerDiakogiannis2020Deep,AungEtAl2020Farm,WangEtAl2022Unlocking,MeiEtAl2022Using,KernerEtAl2023Multi,dua2024agriculturallandscapeunderstandingcountryscale}, which differ from natural (RGB) images in several aspects such as their geometry, atmospheric effects, spectral bands and radiometric information \citep{LillesandKieferChipman2015}.

An important practical constraint is posed by the spatio-temporal tradeoff in this context -- limitations in remote sensing technologies and high hardware costs prohibit us from obtaining simultaneously high temporal resolution and high spatial resolution images on a large scale \cite{Zhu2018Updated,MengEtAl2022Deep}.
E.g., very high resolution (VHR) satellite data (e.g., $\le 1$ m/px Ground Sample Distance (GSD), from Airbus, WorldView satellites) which contains detailed spatial information, essential for fine-grained segmentation of smallholder farms, is only available (bi-)annually; while low resolution (LR) satellite data (e.g., 10-30 m/px GSD from Sentinel-2, Landsat-8, PlanetScope satellites) are available every 1-8 days but lack details necessary for this task.
Since agricultural processes are dynamic and seasonal, field boundaries may change at higher temporal frequency (up to bi-weekly \citep{sapt}, e.g., see \cref{fig:TOF}) and delineating them from LR satellite images is nearly impossible for many smallholder farms.

To address this problem, Reference-based Super-Resolution (Ref-SR) techniques can be used which utilize image super-resolution (SR) methods to obtain HR images from one or more LR images at a given time period, with guidance from previously available co-located HR image.
Reconstructing semantically correct and perceptually plausible HR images via Ref-SR requires (i) adequate semantic and texture modeling of land cover changes between Ref and LR, and (ii) in the satellite context, modeling the differences in sensors used which make the LR and HR images aligned only at a semantic level and not at pixel level.
These issues have been addressed in 
various deep learning models have been developed for Ref-SR
\cite{c2,lu2021masa,huang2022task,dong2021rrsgan,zhang2022rrsr,amsa,zhang2023reference,aslahishahri2023darts,dong2015imagesuperresolutionusingdeep,hsr-diff,dong2024building}.
The desired segmentation is obtained from the super-resolved image (see fig. \ref{fig:TOF}). 

However, three challenges remain, which hinder the effective use of such Ref-SR based approaches for smallholder farm boundary delineation.
First, the scale factor of existing methods remain limited to less than 16$\times$
(see \cref{sec:related work}),
which is insufficient to discern subtle shifts in field boundaries (see fig. \ref{fig:TOF}). 
Since pixel-based SR methods are optimized for perceptual quality and not task-specific accuracy, high-frequency image features related to fine-scale and inter-field boundary integrity  
are lost. 
Second, most SR methods use only RGB inputs, with some using an additional NIR band, and do not effectively utilize multi-spectral, multi-temporal information available from diverse Earth Observation data sources. 
Finally, a limitation found in most previous works is due to the use of the downsampled versions of HR images as `proxies' for real LR images.
In reality, HR and LR images are rarely acquired simultaneously, leading to genuine temporal changes on the ground (e.g., vegetation phenology, new construction, transient objects) rather than  resolution differences only. Furthermore, images from different sensors inherently possess distinct characteristics, creating a `style gap' in terms of color rendition, texture, and radiometric properties. 
Models trained solely on such proxy LR images 
fail to generalize when confronted with the complex degradations present in real-world satellite imagery  \citep{xiao2023d2u}. 

We develop a new method,  SEED-SR (\textbf{S}egmentation \textbf{E}mbedding \textbf{E}nhancement via \textbf{D}iffusion - for \textbf{S}uper \textbf{R}esolution) to address these problems by utilizing a combination of conditional latent diffusion models \citep{rombach2022highresolutionimagesynthesislatent}, and a large-scale multi-spectral, multi-source geo-spatial foundation models (FM),
which are increasingly being developed \cite{dumeur2024paving,brown2024better,astruc2024anysat,lacoste2023geo,jakubik2023foundation,zhu2024foundations,nguyen2023climax}. The key idea of our approach is to bypass the explicit SR task in the pixel space and instead perform SR in a segmentation-aware latent space.
Thus, in SEED-SR we `super-resolve' an LR embedding to an HR embedding, using latent  diffusion models, which offers the benefits of improved fidelity, and scale consistency 
\cite{rombach2022highresolutionimagesynthesislatent}. 
We obtain the embeddings through FMs, thereby reducing our training time and effectively utilizing large-scale, diverse data that is used in pretraining the FMs.
However, the FMs yield fixed-size embeddings which are very high-dimensional which makes diffusion challenging, 
a problem that we solve through architectural innovations within SEED-SR.
Thus, we demonstrate how rich feature hierarchies learned from geo-spatial FMs 
can be effectively utilized 
to solve an open challenge in super resolution.

In summary, this work makes the following key contributions:
\begin{itemize}[noitemsep,topsep=0pt,labelindent=0em,leftmargin=*]
\item We develop SEED-SR, 
a method to generate very high resolution segmentation maps (at 50cm GSD) for field boundary delineation, at an unprecedented 20$\times$ super-resolution, from multi-temporal, low-resolution multi-source, multi-spectral satellite image inputs (at 10m GSD) and a high resolution historical reference image.

\item SEED-SR showcases a unique way to leverage multiple pre-trained large-scale geo-spatial foundation models, and to learn a mapping between their high-dimensional embeddings with latent diffusion models.

\item We rigorously evaluate our method on real-world data comprising LR and HR images from different satellite sensors.
Our experiments on the field boundary delineation task, for smallholder farms across multiple countries in the Global South, show that our method outperforms state-of-the-art methods following the two-step pixel-level SR and segmentation approach.  

\end{itemize}

\section{Related Work}
\label{sec:related work}

\textbf{Super-resolution} (SR) has been extensively studied in computer vision \cite{su2024review,liu2024arbitrary,moser2024diffusion} and in remote sensing  (RS)
\cite{sdraka2022deep}.
We summarize the closest related works in Table \ref{tab:sr_comparison_simple} along 6 dimensions -- whether or not the method (1) was designed for RS images \textbf{[RS]}, (2) was trained/optimized for segmentation \textbf{[OPT]}, (3)  was designed/evaluated with real LR satellite data \textbf{[Real]},  (4) whether the multispectral LR input was used \textbf{[MS]} (5) considered Ref images  as inputs \textbf{[Ref]}, (6) their maximum scale factor reported \textbf{[SF]}.
\begin{table}[htbp]
\captionsetup{font=small}
\centering
\footnotesize
\caption{Summary of Related Works}
\label{tab:sr_comparison_simple}
\begin{tabular}{
    >{\arraybackslash}p{2.25cm} 
    >{\centering\arraybackslash}p{0.5cm} 
    >{\centering\arraybackslash}p{0.5cm} 
    >{\centering\arraybackslash}p{0.5cm} 
    >{\centering\arraybackslash}p{0.5cm} 
    >{\centering\arraybackslash}p{0.5cm} 
    >{\centering\arraybackslash}p{1cm} 
}
\toprule
\textbf{Method} & \textbf{RS} & \textbf{OPT} & \textbf{Real} & \textbf{MS} &  \textbf{Ref} & \textbf{SF}\\
\midrule
\citep{dong2015imagesuperresolutionusingdeep,nguyen2022singleimagesuperresolutiondual, scaleaware}  & \xmark & \xmark & NA & \xmark & \xmark & 4x \\
\citep{funsr,7937881,9654169, wu2024latentdiffusionimplicitamplification} & \cmark & \xmark & \xmark & \xmark  & \xmark & 6x \\
\citep{ttsr,wtrn,c2,amsa,datsr} & \xmark & \xmark & NA & \xmark & \xmark & 8x  \\
\citep{lu2021masa,huang2022task,zhang2022rrsr,aslahishahri2023darts} & \xmark & \xmark & NA & \xmark &  \cmark & 8x \\
\citep{dong2021rrsgan,zhang2023reference,dtesr} & \cmark & \xmark & \xmark & \xmark  & \cmark & 4x\\
\citep{dong2024building,dcdmf}& \cmark & \xmark & \xmark & \xmark &  \cmark  & 16x  \\
\citep{wang2024semanticguidedlargescale} & \cmark & \xmark & \cmark & \xmark & \xmark  & 16x\\
\citep{Zabalza2022SuperResolution,10414975,10887321,8897860,Galar2020SuperResolution} & \cmark & \xmark & \xmark & \cmark  & \xmark & 4x \\
\citep{dff,segesrgan} & \cmark & \cmark & \cmark & \xmark  & \xmark & 3x-5x \\
\textbf{SEED-SR (Ours)} & \cmark & \cmark & \cmark & \cmark & \cmark & \textbf{20x} \\
\bottomrule
\\
\end{tabular}
\end{table}

Many SR techniques have been developed based on deep learning architectures 
for both general settings e.g., 
SRCNN \citep{dong2015imagesuperresolutionusingdeep}, DINN \citep{nguyen2022singleimagesuperresolutiondual}, and SADN\citep{scaleaware}
and specifically for remote sensing (RS), e.g.,
FunSR\citep{funsr}, LGCNet \citep{7937881}, TransENet \citep{9654169}.
Generative Adversarial Networks (GAN) 
have been widely used in SR for remote sensing applications (e.g., TTSR \citep{ttsr}, WTRN \citep{wtrn}, C2-Matching \citep{c2},
AMSA \citep{amsa}, and DATSR \citep{datsr}).
However, GANs 
have largely been outperformed by diffusion-based models \cite{saharia2021imagesuperresolutioniterativerefinement} which offer more realistic reconstruction and stable training. 

Very few SR techniques have utilized multi-spectral LR inputs, and they usually just use the NIR band in addition to RGB, e.g., in\citep{Zabalza2022SuperResolution,10414975,10887321,8897860,Galar2020SuperResolution}. Experimental comparisions in \citep{10494508} against TransENet \citep{9654169} show that addition of NIR does not significantly improve performance over RGB. Most focus on the SR task only and not on downstream analysis after SR, notable exceptions include
\citep{segesrgan,dff} who use multi-task networks for joint segmentation and super-resolution.
None of these use Ref images to guide SR, and most achieve low scale factors of $\le$ 8x.

\noindent
\textbf{Reference-based SR}:
Among SR methods, our work belongs to the category of Ref-SR models \cite{c2,lu2021masa,huang2022task,dong2021rrsgan,zhang2022rrsr,amsa,zhang2023reference,aslahishahri2023darts}. 
Most recent Ref-SR models for remote sensing have been obtained through conditional diffusion, e.g.,  DCDMF \cite{dcdmf} and HSR-Diff \citep{hsr-diff} and RefDiff \citep{dong2024building}, which incorporate various conditions derived from the LR and Ref images into the denoising process of diffusion.
The state-of-the-art RefDiff model
improves texture reconstruction
through the use of land cover change priors (obtained via change detection algorithms) and the use of SFT layers \cite{sft} which enable semantics-guided
 and reference texture-guided denoising.
However, in terms of addressing the problem of large (20x) spatial resolution gap, its performance remains limited, also seen in our experiments.

Further, there are two limitations found in most previous works on SR for remote sensing.
First, as discussed earlier, proxy LR images from downsampled HR images lead to models that are not generalizable. 
Second, as shown in previous works (e.g., \citep{wang2024semanticguidedlargescale}) and our experiments, enhancing general visual quality through SR does not directly translate to the high accuracy required for downstream applications.
Since image-centric SR typically optimizes for global perceptual quality or pixel-wise fidelity (e.g., PSNR/SSIM), they can inadvertently smooth over or misrepresent the subtle yet crucial features necessary for fine-grained segmentation.
In our work, we address both these limitations. 

\noindent
\textbf{Farm boundary delineation}:
Previous deep learning approaches developed specifically for smallholder field boundary segmentation, such as  
\cite{WangEtAl2022Unlocking,MeiEtAl2022Using,dua2024agriculturallandscapeunderstandingcountryscale},
have utilized HR images only and have not tackled the SR problem; and 
have also not used multi-spectral, multi-source inputs.
\section{Problem Statement}
Let $X_{t,h}$ denote a HR, RGB image at time $t$ and 
$S_{t,h}$ denote a segmentation map obtained from $X_{t,h}$ and let $X_{t',r}$ denote a single-source HR, RGB
Ref image  from the same geographical area, captured at time  $t' < t$ with at least a 6-month interval between $t'$ and $t$. We specifically select a 6 months interval to ensure that there are noticeable landscape differences between $X_{t,h}$ and $X_{t',r}$.
Let $[X_{t_m,l}, \dots, X_{t,l}]$ represent a sequence of  multi-temporal, multi-spectral, multi-source LR 
images captured within a temporal window of size $t-t_m$ such that $t' \leq t_m < t$. Our aim is to develop a model $\mathcal{M}: ([X_{t_m,l}, \dots, X_{t,l}], X_{t',r}) \rightarrow S_{t,h}$ which learns a segmentation map, for field boundary delineation, from the input LR and Ref images.

\section{Our Approach: SEED-SR}
\begin{figure*}
  \centering
    \captionsetup{font=footnotesize}
  \includegraphics[width=0.9\linewidth]{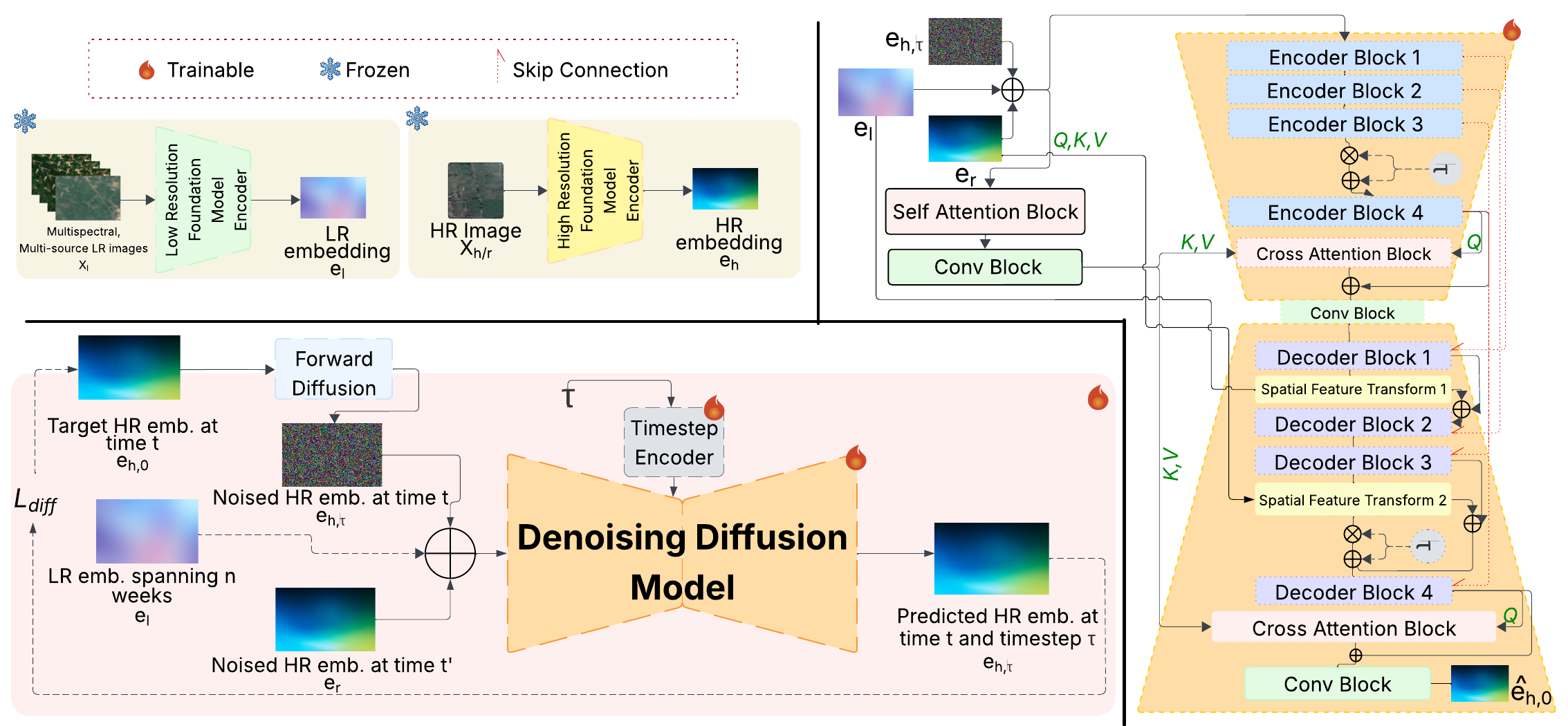}
  \caption{(Left) Training Diagram: A conditional latent diffusion U-Net predicts a clean HR embedding ($e_{h,0}$) from a noised input ($e_h$), conditioned on concatenated LR ($e_l$) and reference HR ($e_r$) features. (Right) Architecture of the conditional denoising diffusion model($G_{\theta}$). Best viewed in colour.}
  \label{fig:model_training}
  \vspace{-0.4cm}
\end{figure*}

We build our desired model $\mathcal{M}$ using three key components: two distinct Geo-spatial Foundation Models (FM) and a conditional diffusion model. 
The first FM is the AlphaEarth Foundations model \cite{brown2024better} a pre-trained model to obtain task-agnostic features from multi-temporal, multi-spectral, multi-source LR images. 
This allows us to obtain embeddings, $e_l$, of our LR input sequence $[X_{t_m,l}, \dots, X_{t,l}]$.
The second FM is an encoder-decoder model \citep{SchottlanderShekel2025Geospatial} which takes HR images as input and is specifically fine-tuned for agricultural field segmentation.
We use the encoder derived from this FM to obtain embeddings, $e_r, e_h$ of Ref ($X_{t_m,r}$) and HR ($X_{t,h}$) images respectively.
Thus, through the use of both the FMs, our problem boils down to learning the distribution of HR embeddings ($e_h$) 
from the LR embeddings ($e_l$) and the Ref embeddings ($e_r$). 
This is challenging because of the high dimensionality of the latent information-rich embeddings, $e_r, e_h$.
We develop a novel adaptation of the conditional denoising diffusion model \citep{ddpm} to learn $e_h$, conditioned on $e_r, e_l$.
With the learnt HR embeddings, the decoder of the second FM can be used to obtain the required segmentation map $S_{t,h}$.
Fig. \ref{fig:model_training} shows the overall schematic.

\subsection{Foundation Models for Geospatial Reasoning}
We utilize two distinct pre-trained foundation models:
\begin{itemize}[noitemsep,topsep=0pt,labelindent=0em,leftmargin=*]

     \item Low Resolution Foundation Model (LR-FM): We use the pre-trained AlphaEarth Foundations Model \citep{brown2024better} for processing LR imagery. The model utilizes multi-spectral images from multiple sources for a given time period ($[t_m \ldots t]$) -- Sentinel-2 L1C top-of-atmosphere (bands B2,3,4,8,11), Landsat 8/9 T1 TOA (bands B2,3,4,5,6,8,10) and Sentinel-1 GRD (bands HH, HV, VV, VH depending on availability), all resampled to 10m and summarizes them into embeddings of size (128,128,64) pixels spanning an area of 1280 $\times$ 1280 $m^2$. We generate embeddings that span the 4 weeks before time t. We denote the pre-trained encoder derived from the model by $E_{LFM}$.
     
    \item High Resolution Foundation Model (HR-FM): This encoder-decoder model \citep{SchottlanderShekel2025Geospatial}, is pre-trained and fine-tuned for agricultural field segmentation specifically for smallholder farms. The model processes HR images (here, 0.5 m GSD covering 320 $\times$ 320 $m^2$, resized to 480 $\times$ 480 pixels for the encoder).
    The encoder, $E_{HSM}$, produces (120,120,3840) dimensional feature maps. 
    The decoder, $D_{HSM}$, 
    is used to obtain the final HR segmentation map.


\end{itemize}

We use encoders of these FMs to obtain the embeddings: $e_l = E_{LFM}([X_{t_m,l}, \dots, X_{t,l}]), \: 
e_h = E_{HSM}(X_{t,h}), \: e_r = E_{HSM}(X_{t',r})$.
We sample an area of 320 $\times$ 320 $m^2$ from the model output of $E_{LFM}$ corresponding to the area in the HR images to yield a (32,32,64) sized tensor. 
To integrate $e_l$ with spatial feature maps $e_r$ and $e_{h}$ this 64D vector is spatially tiled and projected to match the 
dimensions  of $e_r$.





\subsection{Latent Diffusion for Embedding Generation}
We estimate
the desired conditional distribution,
$p(e_h | e_l, e_r)$ by training a parameterized model, $G_{\theta}$.
To this end, we adapt the denoising diffusion probabilistic model (DDPM) \citep{ddpm} to operate in the latent space of $E_{HSM}$, i.e., to predict the embedding $e_h$ from its noise-perturbed version. 
Following standard DDPM, our diffusion model consists of a forward process which gradually adds Gaussian noise to the HR embedding $e_h$ over T discrete timesteps following a predefined noise variance schedule and a reverse process which learns to recover the original embedding $e _{h,0}$ from its noised($\epsilon$) versions $e_{h,\tau}$ conditioned on $e _l, e_r$ and the timestep $\tau$ where $\tau \epsilon [1,T]$. 
This is achieved by training a neural network $G_{\theta}(e_{h,\tau}, \tau, c)$ to predict ${e}_{h,0}$ , where $c=(e_l,e_r)$ represents the combined conditioning information. 
The model is trained by minimizing the mean squared error between the true embedding and its prediction: $L_{\text{DDPM}} = \mathbb{E}_{\tau, e_{h,0}, \varepsilon} \left[ || e_{h,0} - G_{\theta}(e_{h,\tau}, \tau, c) ||^2_2 \right]$.

\subsection{Handling High-Dimensional Embeddings}
\label{dims}
Our network $G_{\theta}(e_{h,\tau}, \tau, c)$ follows a U-Net architecture \citep{ronneberger}, with novel architectural elements, strategically designed to extract and utilize relevant information at each stage, from high-dimensional inputs. 
Appendix \ref{app:model_details} has details of each block and dimensions of all the layers.

\subsubsection{Encoder Design}

Our encoder consists of 4 encoder blocks, each comprising a convolution block (as detailed in Appendix \ref{app:model_details}) and a max pooling layer.
It processes a noised HR (120,120,3840)-size embedding as its primary input, along with a reference HR (120,120,3840)-size embedding, a spatially resized LR (32,32,64)-size embedding, and a diffusion timestep identifier.
The encoder path progressively spatially downsamples 
this combined input, integrating the timestep embedding.
However, the channel depth is progressively increased  (320, 480, 560, 640, and a bottleneck of 784 channels). These substantial channel capacities are deliberately chosen to ensure that the rich information contained within the initial high-dimensional embeddings is effectively propagated and transformed without significant loss.


{\bf Self Attention, Convolution and Cross Attention.}
Initially, the input modalities (upsampled $e_l$ , $e_r$ and the noised high-resolution input $e_h$) are concatenated, forming a high-dimensional tensor with 7744 channels. This concatenated input undergoes a self-attention mechanism. This crucial step allows the model to learn complex spatial correlations and inter-dependencies across the different input features. Following self-attention, the resulting feature map is a (120, 120, 64) dimensional tensor, effectively condensing the information while preserving key spatial relationships.
To integrate this contextual information with the model's encoder pathway, this (120, 120, 64) attention output is spatially downsampled (to 7x7). The downsampling reduces memory usage from ~14 GB to ~2.75 MB and FLOPs from 423 GFLOPs to 5 MFLOPs making diffusion feasible in this feature space. Additional details about memory fitting are in \cref{app:novel} . This downsampling is achieved using a convolutional block (comprising of 2D convolutions, LayerNorm(s), and ReLU activations (as detailed in Appendix \ref{app:model_details}) to match the spatial dimensions (7x7) of the features at the deepest part of the encoder (Encoder Block 4). This downsampled attention map, rich in global context, is then fused via cross-attention with the output of the encoder block 4 (which has 640 channels). This allows the encoder's features to be conditioned by the global context derived from all input modalities.
Furthermore, the decoder also leverages this contextual understanding in its final stage via cross-attention. 






\subsubsection{Decoder Design}

Each of the 4 decoder blocks comprises transposed convolution and convolutional layers.
The decoder path reconstructs the spatial resolution using skip connections from the encoder, and incorporates the timestep embedding. 
Inspired by the approach in \citep{dong2024building}, we strategically apply
conditioning using Spatially-adaptive Feature Transform (SFT) blocks \cite{sft}
in the decoder.
Each SFT layer encodes the function $F_{out} = \gamma \odot F_{in} + \beta$ where $F_{out}, F_{in}$ are the output and input features of the layer and the scale ($\gamma$) and shift ($\beta$) parameters are adaptively learnt from the input conditions (see Appendix \ref{app:model_details} for more details).
The resized LR embedding modulates early-stage decoder features using a a SFT block, while the reference HR embedding similarly conditions mid-stage decoder features through another SFT block.
SFT layers play an important role in our decoder as they allow the $e_l$ and $e_r$ embeddings to provide detailed, spatially varying, and nuanced conditioning signals. 

Before the final output, late-stage decoder features are further refined by a cross-attention mechanism employing the fused HR contextual representation established at the input stage, enabling the decoder to refine its predictions using this comprehensive, early-stage feature summary. 
The network culminates in a convolutional layer that outputs a direct prediction of the input noised HR embedding.

\begin{figure*}
  \centering
    \captionsetup{font=footnotesize}
  \includegraphics[width=\linewidth]{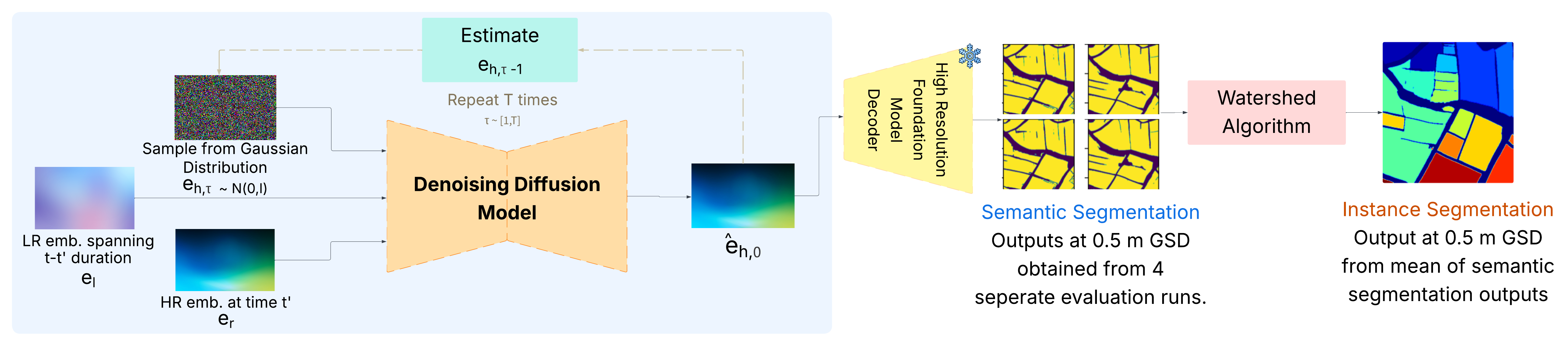}
  \caption{Inference uses a denoising diffusion framework: the model iteratively estimates the clean HR embedding ($\hat{e}_{h,0}$) from a noised $e_{h,\tau}$, conditioned on LR ($e_l$) and Ref ($e_r$). This final $\hat{e}_{h,0}$yields a semantic segmentation map via a frozen $D_{HSM}$ decoder. 4 samples are generated from distinct noise seeds; their semantic segmentation logits are averaged and then watershed-processed for instance segmentation. Best viewed in colour.
 } 
  \label{fig:model_inference}
  \vspace{-0.4cm}
\end{figure*}


\subsection{Segmentation Output Generation}
During inference, 
the pretrained FM encoders ($E_{LFM}, E_{HSM}$) are used to obtain
the embeddings $e_l, e_r$, 
which, in turn, are used in the trained denoising diffusion model along with a sampled latent from an Isotropic Gaussian distribution $\mathcal{N} \sim (0,I)$ to obtain the predicted HR embedding  $\hat{e}_{h,0}$.
This embedding is used in the pretrained decoder $D_{HSM}$
of HR-FM to obtain a semantic segmentation map. In order to promote sample variation, we use four different seeds for different inference runs and compute the mean of semantic segmentation logits obtained from the runs. The mean of semantic segmentation logits are fed to a watershed algorithm \citep{watershed} to obtain an instance segmentation map. 
\section{Experimental Evaluation}
\subsection{Datasets and Preprocessing}
\label{sec:dataset}
Our experiments are on two datasets.
The first comprises images from {\bf Vietnam}, taken from a publicly available benchmark dataset, AI4SmallFarms \citep{ai4}, for agricultural field boundary delineation.
The images are collected between 2021 and 2023, and the train/val/test splits contain 5296/432/753 images.
The second is a larger proprietary dataset spanning the entirety of {\bf India}, collected between 2019-22, with 32611/291/2615 images in train/val/test splits.
In both cases, the satellite images encompass smallholder farms at national scales across diverse landscapes
including farms on hilly terrains, arid regions, river plains and other areas.
The high-resolution images are processed into
(640,640) tiles at a 0.5m GSD,  co-located low-resolution embedding are obtained from $E_{LFM}$ at 10m GSD. 
Embeddings are generated using the $E _{LFM}$ encoder, initially producing (128,128,64) feature maps which are subsequently cropped to (32,32,64) to correspond with the high-resolution image areas. Thus both HR images and $e_l$ are co-located and represent the same 320*320 $m^2$ area. Additional details about the dataset collection and labelling are present in Appendix \ref{app:expts}, with a summary of input details in Table \ref{app:input_dets}.
\subsection{Experiment Settings}
To rigorously validate our approach, we perform evaluation against two complementary sets of reference annotations. The first, termed `\textbf{teacher predictions}', represents the HR-FM's inference on the target high-resolution imagery. The second comprises `\textbf{human labels}', which serve as our manually curated ground truth. Against these human labels, the teacher predictions demonstrated significant agreement (see Appendix \ref{app:expts}).
Also, results on teacher predictions allow us to compare with the case of segmentation directly on an HR image, if it were available at the same time.
We use semantic and instance segmentation metrics ($mIoU_S, mIoU_I$) and pixel-level binary classification metrics to evaluate performance.
Formal definitions are in Appendix \ref{app:expts}.
Appendix \ref{app:model_details} has hyperparameter details.
\subsection{Baseline Methods}
Our baselines follow the common two-step approach 
wherein SR, in the pixel space, is performed first and the super-resolved image is then segmented.
We use 4 categories of SR methods for benchmarking
(i) Bicubic upsampling, serving as a standard interpolation baseline; (ii) competitive SR methods on natural images: SRCNN \citep{dong2015imagesuperresolutionusingdeep}, DINN \citep{nguyen2022singleimagesuperresolutiondual}; (iii) SR algorithms specifically designed for remote sensing imagery: FunSR\citep{funsr}, LGCNet\citep{7937881} and (iv) the state-of-the-art Ref-SR method
RefDiff\citep{dong2024building}, which was shown to outperform
GAN-based methods, 
TTSR \citep{ttsr}, WTRN \citep{wtrn}, C2-Matching \citep{c2},
AMSA \citep{amsa}, and DATSR \citep{datsr} as well as
diffusion-based models,
EDM \cite{karras2022elucidating} and HSR-Diff \citep{hsr-diff}.
We feed changes in field boundary segmentation masks as the change prior to RefDiff. 
Since baseline methods cannot utilize our multi-spectral multi-source inputs, we follow their settings closely, i.e., use downsampled HR images as LR `proxies' and the recommended hyperparameters.
We use HR-FM to obtain segmentation maps from the super-resolved images for all baselines, thus ensuring a fair comparison with our method which uses the decoder, $D_{HSM}$, of HR-FM.

\subsection{Results: Field Boundary Segmentation}

Table \ref{tab:segmentation-comparison-our-dataset} shows the results of our method and all the baseline methods. 
\ref{app:expts} details the evaluation metrics utilized for comparision.
In both the datasets we observe the same performance trend: 
Among the baselines, methods specifically designed for satellite images, FunSR, LGCNet and RefDiff, outperform the rest; and RefDiff which effectively utilizes a Ref image in its modeling performs the best.
All the baselines, where pixel-based SR is followed by segmentation, are outperformed by our method across all the metrics.
In particular,  our model achieves 25.5\%  and 14.67\%  improvement in $mIoU_I$  
and 
12.9\% and 10.87\% 
improvement in $mIoU_S$  
 over RefDiff 
in the two datasets respectively.

\begin{table*}[htbp] 
\captionsetup{font=footnotesize}
  \caption{Segmentation Performance on India (above) and Vietnam (below) datasets. Comparison with teacher predictions (left) and  human labels (right). Metrics include -- A: Accuracy, P: Precision, R: Recall, F1: F1 Score and mean $IoU$ for instance ($_I$) and semantic ($_S$) segmentation.}
\label{tab:segmentation-comparison-our-dataset} 
\centering
\scriptsize
\begin{tabular}{lcccccccccccc} 
\toprule
& \multicolumn{12}{c}{\textbf{India} Dataset} \\ 
\cmidrule(lr){2-13} 
& \multicolumn{6}{c}{\textbf{Teacher Predictions}} & \multicolumn{6}{c}{\textbf{Human Labels}} \\ 
\cmidrule(lr){2-7} \cmidrule(lr){8-13} 
Method & $mIoU_{I}$ & $mIoU_{S}$ & $A$ & $P$ & $R$ & $F1$ & $mIoU_{I}$ & $mIoU_{S}$ & $A$ & $P$ & $R$ & $F1$ \\ 
\midrule
Bicubic & 10.9 & 15.25 & 59.41 & 16.21 & 8.07 & 8.84 & 10.3 & 15.22 & 60.7 & 15.66 & 8.66 & 9.08 \\
SRCNN & 28.71 & 37.25 & 69.91 & 36.92 & 23.67 & 24.57 & 27.23 & 35.81 & 68.10 & 36.86 & 21.63 & 23.32 \\
DINN & 31.29 & 43.92 & 71.03 & 38.47 & 24.94 & 25.37 & 28.85 & 40.16  & 69.41 & 37.97 & 23.12 & 25.64 \\
LGCNet & 32.05 & 42.96 & 71.15 & 40.13 & 28.83 & 28.10 & 30.76 & 41.29 & 69.03 & 39.25 & 24.19 & 26.42 \\
FunSR & 37.41 & 48.27 & 74.59 & 44.32 & 32.00 & 32.40 & 35.15 & 45.07 & 71.78 & 43.62 & 28.82 & 30.21 \\
RefDiff & 47.96 & 65.73 & 81.48 & 58.02 & 54.77 & 54.05 & 45.10  & 62.94 & 79.18 & 57.11 & 50.59 & 51.59 \\
\textbf{SEED-SR(Ours)} & \textbf{58.44}  & \textbf{74.96} & \textbf{86.33} & \textbf{62.11} & \textbf{64.68} & \textbf{61.51} & \textbf{56.61} & \textbf{71.06} & \textbf{83.55} & \textbf{60.94} & \textbf{59.38} & \textbf{58.34} \\ 
\bottomrule
\\
  \end{tabular}

  \begin{tabular}{lcccccccccccc} 
    & \multicolumn{12}{c}{\textbf{Vietnam} Dataset (AI4SmallFarms)} \\ 
\cmidrule(lr){2-7} \cmidrule(lr){8-13} 
    Method & $mIoU_{I}$ & $mIoU_{S}$ & $A$ & $P$ & $R$ & $F1$ & $mIoU_{I}$ & $mIoU_{S}$ & $A$ & $P$ & $R$ & $F1$ \\ 
    \midrule
    Bicubic & 15.70 & 24.48 & 61.67 & 29.47 & 19.37 & 18.83 & 14.67 & 23.32 & 56.28 & 32.41 & 18.32 & 18.28 \\
    SRCNN  & 28.23 & 40.92 & 69.59 & 49.06 & 34.44 & 35.61 & 27.73 & 38.07 & 68.32 & 47.98 & 31.86 & 32.15\\
    DINN & 32.71 & 45.64 & 71.09 & 53.91 & 37.18 & 39.14 &29.92 & 44.45 &69.55 &52.86 & 35.31 & 39.08\\
    LGCNet & 35.96 & 48.61 & 71.28 & 58.17& 37.61& 43.58 & 32.59 & 47.03 & 68.19 & 57.40 &34.78 &41.24  \\
    FunSR  & 37.16 & 56.24 & 72.85 & 62.06 & 41.18 & 47.96 & 35.59 & 54.23 & 69.52 & 59.02 & 40.76 & 55.44 \\
    RefDiff & 44.82 & 63.02 & 77.43 & 71.49 & 66.32 & 68.24 & 43.16 & 60.98 & 75.17 & 69.56 & 63.51 & 67.39 \\
    \textbf{SEED-SR(Ours)}  & \textbf{50.91}  & \textbf{71.01} & \textbf{81.34} & \textbf{76.76} & \textbf{74.97} & \textbf{74.43} & \textbf{49.49} & \textbf{67.61} & \textbf{78.72} & \textbf{72.94} & \textbf{75.65} & \textbf{71.89}    \\ 
    \bottomrule
      \\
  \end{tabular}
\vspace{-0.7cm}
\end{table*}

\subsection{Results: Qualitative Comparison}
\begin{figure*}
  \centering
    \captionsetup{font=footnotesize}
  \includegraphics[width=\linewidth]{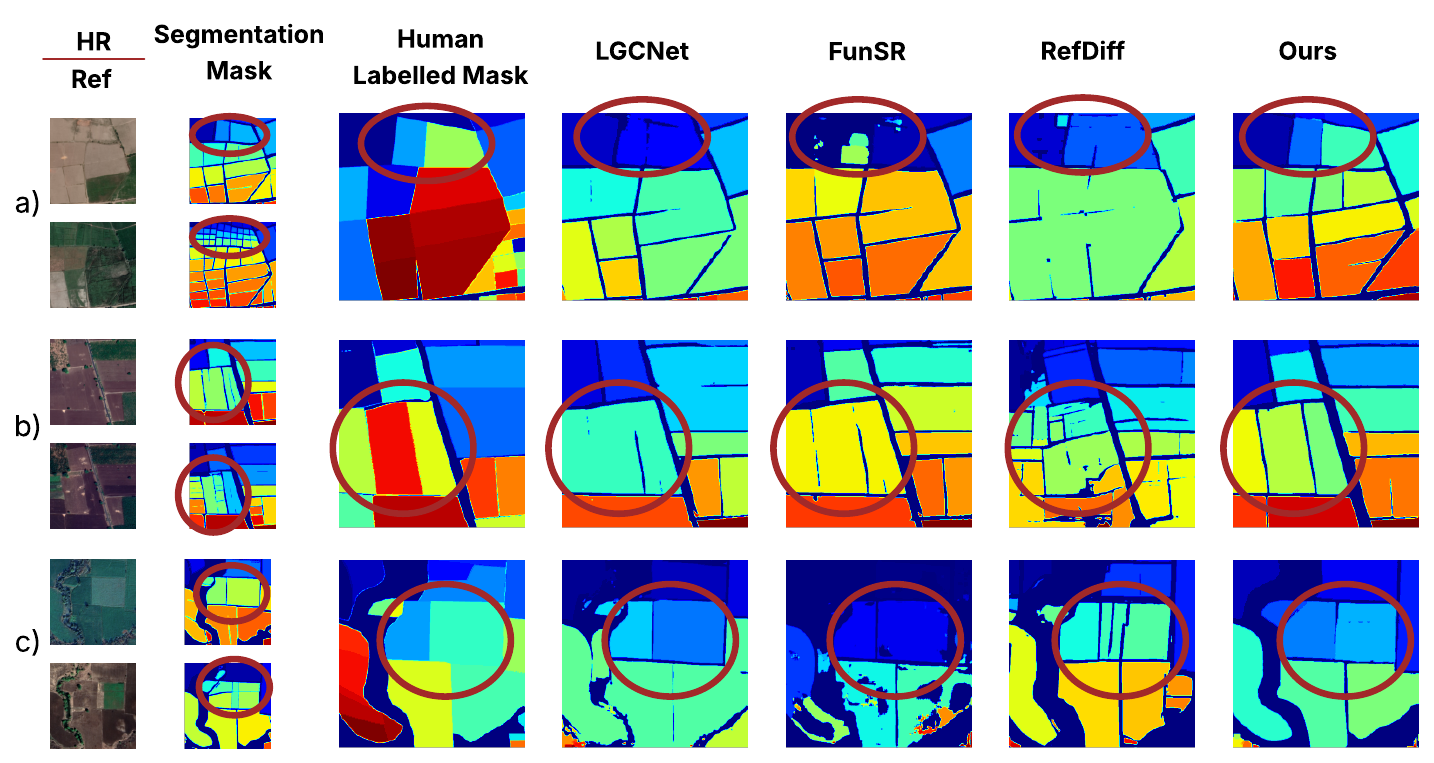}
  \caption{Results on 3 examples from the India Dataset. Best viewed in colour.}
    \label{fig:results}
    \vspace{-0.3cm}
\end{figure*}

Fig. \ref{fig:results} illustrates the output masks obtained by our method and 3 baselines for a few sample inputs.
The first two columns show the Ref and HR images and the their segmentation masks as predicted by HR-FM.
The circled portions of the masks indicate the changes which have occurred in the field boundaries -- in the first and second row we observe that multiple farms have merged into a larger farm. In both cases, our method delineates the boundaries more accurately compared to the baselines.
In the last row we observe the noisy artifacts added during SR by other methods, which are absent in our reconstruction. 
More examples are discussed in Appendix \ref{app:expts}, including cases where our method does not perform well.

\subsection{Results: Geographic Generalizability}
To quantitatively demonstrate geographical generalization, we conducted zero-shot generalization experiments. We used our model, trained only on the India dataset \cref{sec:dataset}, and evaluated its performance on two unseen regions from different continents:
(i) \textbf{Vietnam} and (ii)
\textbf{Kenya}.
The former is the same AI4SmallFarms \citep{ai4} test set described in \cref{sec:dataset}, while the latter is from \citep{KernerEtAl2023Multi}.
On Vietnam, our India-trained model achieves \textbf{46.24} $mIoU_I$ against human labels and \textbf{47.62} $mIoU_I$ against teacher predictions demonstrating strong generalization across different smallholder farming systems in Asia. Notably, the India-trained model’s performance on Vietnam (46.24 $mIoU_I$) exceeds the best baseline (RefDiff 44.82 \citep{dong2024building}) even when RefDiff was fully trained on the Vietnam data (see \cref{tab:segmentation-comparison-our-dataset}).
On Kenya, the India-trained model achieves \textbf{51.49} $mIoU_I$ against human labels and \textbf{53.42} $mIoU_I$ against teacher predictions, demonstrating strong generalization across farming systems in a different continent.

\begin{table}[htbp]
\centering
\footnotesize
\caption{Zero shot performance ($mIoU_I$) across geographies.} 
\label{tab:zeroshot}
\begin{tabular}{
    >{\centering\arraybackslash}p{1cm} 
    >{\centering\arraybackslash}p{2cm} 
    >{\centering\arraybackslash}p{2.5cm}   
}
\toprule
\textbf{Dataset} & \textbf{Human Labels} & \textbf{Teacher Predictions} \\
\midrule
\textbf{Vietnam} & 46.24 & 47.62 \\
\textbf{Kenya} & 51.49 & 53.42 \\
\bottomrule
\end{tabular}
\end{table}

\subsection{Results: Ablation Studies on SEED-SR}
\begin{table}
  \centering
  \captionsetup{font=footnotesize}
  \caption{Ablation Studies}
  \label{tab:ablation}
  \footnotesize
  \begin{tabular}{
    p{1.4cm} 
    >{\centering\arraybackslash}p{0.5cm} 
    >{\centering\arraybackslash}p{0.5cm} 
    >{\centering\arraybackslash}p{0.4cm} 
    >{\centering\arraybackslash}p{0.4cm} 
    >{\centering\arraybackslash}p{0.5cm} 
    >{\centering\arraybackslash}p{0.5cm} 
  }
    \toprule
    & \multicolumn{2}{c}{Components} & \multicolumn{2}{c}{Model Inputs} & \multicolumn{2}{c}{Teacher Predictions} \\
    \cmidrule(lr){2-3} \cmidrule(lr){4-5} \cmidrule(lr){6-7}
    Method Config & $E_{LFM}$ & $E_{HSM}$ & LR & Ref & $mIoU_I$ & $mIoU_S$ \\
    \midrule
    A  & \xmark & \xmark & \cmark & \cmark & 37.41 & 41.78 \\
    B  & \xmark & \xmark & \cmark & \xmark & 16.27 & 23.41 \\
    C  & \cmark & \xmark & \cmark & \xmark & 17.49 & 25.83 \\
    D  & \xmark & \cmark & \xmark & \cmark & 48.73 & 56.19 \\
    E  & \xmark & \cmark & \cmark & \cmark & 52.19 & 65.28 \\
    \textbf{SEED-SR} & \cmark & \cmark & \cmark & \cmark & \textbf{58.44} & \textbf{74.96} \\
    \bottomrule
  \end{tabular}
  \vspace{-0.5cm}
\end{table}
To systematically dissect the impact of $E_{LFM}$, $E_{HSM}$, 
LR and Ref
inputs on the final segmentation performance, as measured by $mIoU_{I}$ and $mIoU_{S}$ on teacher predictions, we conduct ablation studies on the India dataset. 

Table \ref{tab:ablation} shows the results.
In \textbf{Config A} LR and Ref image inputs are utilized without the specialized LFM and HSM encoders i.e., the diffusion model takes the multi-spectral, multi-temporal, Sentinel-2 LR images as input.
Segmentation is performed by using the HR-FM model on the superresolved image (obtained by our diffusion model). 
This configuration yields significantly lower scores 
$(mIoU_I = 37.41, mIoU_S = 41.78)$ compared to our full model. 
This corroborates our 
performance comparison 
in Table
\ref{tab:segmentation-comparison-our-dataset}.
To evaluate the scenario with minimal inputs and no specialized FM-based encoders, \textbf{Config B} utilized only the LR image input without any FM encoders or Ref. The LR input image here was a stack of 32 Sentinel-2 multi-spectral images corresponding to time $\epsilon$ [t',t]. The target was the superresolved $e_{h}$. As expected, this configuration performed poorly further emphasizing the necessity of latent space (via FMs) and the guiding information from the ref HR imagery.

Next, we investigated the importance of each input modality used as conditions in our diffusion model.
In \textbf{Config C}, we removed the HSM encoder for conditioning and the Ref image input, relying solely on the LFM encoder and the LR image sequence. This led to a drastic reduction in performance which clearly demonstrates the critical role of the HR Ref image and its HSM-derived embedding in providing high-fidelity prior information for the diffusion model.
Similarly, \textbf{Config D} isolates the contribution of the Ref image by removing the LFM encoder and the LR image input. While this configuration performs better than relying only on the LR embeddings (Config C), it still performs significantly worse than the proposed method. This underscores that the temporal context and features extracted by the LFM from the LR image sequence provide essential complementary information that the Ref image alone cannot supply. 
Finally, in \textbf{Config E}, a stack of 32 Sentinel-2 LR images corresponding to time $\epsilon$ [t',t] are fed directly while  reference image inputs are encoded with the HSM encoder. This suggests that the specialized feature extraction and representation provided by the LFM encoder for the LR data are vital for optimal performance, reinforcing the benefit of our dual-encoder FM strategy.
The results from Configs C, D and E together strongly indicate that our model learns from both LR and Ref embeddings.
Details of input dimensions and layer parameters across ablation configurations are in Appendix \ref{app:model_details}.
These results demonstrate the importance of each model component and validates that for combined SR and segmentation, a task-specific latent space powered by FMs outperforms pixel-space methods.

\subsection{Results: Additional Experiments }
\label{res:expts}

Appendix \ref{app:expts} has additional results.
To summarize, SEED-SR outperforms two-stage approaches when \textbf{SAM}~\citep{kirillov2023segment}) is used on super-resolved images from various baselines.
\textbf{Sensitivity analysis} shows that (i) SEED-SR's performance depends considerably on the underlying FM -- while diffusion within SEED-SR learns accurate mappings between embeddings, it does not correct for errors in the input embeddings and (ii) clarifies our choice of 4 runs during inference.
\textbf{Running time} analysis of SEED-SR with baselines shows that it is faster than other diffusion-based models (RefDiff), while being slower than non-generative alternatives (FunSR, LGCNet).

\section{Conclusions, Limitations and Future Work}

We develop SEED-SR, a novel latent diffusion-based 
method to generate HR segmentation maps
and demonstrate its 
efficacy 
through experiments on smallholder 
field boundary delineation, 
where it distinguishes itself by 
effectively bridging a 20x resolution gap.
To our knowledge, this work is the first to utilize a combination of latent diffusion model and two 
distinct FMs -- one for processing multi-source, multi-spectral multi-temporal LR inputs and another, 
for defining a task-aware latent space and 
generating HR segmentation maps --
for field boundary delineation.

While our novel framework shows substantial improvements in segmentation performance, there are several limitations which offer avenues for further research.
Our inference is computationally intensive and new techniques to improve its running time could be investigated.
Our method's performance is inherently tied to the underlying FMs, with possibly dissimilar architectures, and achieving optimal semantic and pixel alignment between LR and HR embeddings remains challenging. Ways to  improve the alignment, both through improved FM design and connector architectures such as ours, could be explored. 
Our framework
can easily be adapted to other tasks like object detection by using suitable task-specific FMs and could be evaluated in future work. 
More broadly, our work lays the foundation to build models which can derive fine-grained artifacts reliant on HR details using high-revisit-frequency  LR satellite images to develop more accurate remote sensing applications.

{
    \small
    \bibliographystyle{ieeenat_fullname}
    \bibliography{main}
}
\newpage

\appendix

\section{Societal Impact}
This research stands to make significant positive contributions, particularly in enhancing global food security and supporting sustainable agricultural practices. The developed super-resolution technology specifically addresses the challenge of monitoring agricultural land with high accuracy and timeliness, which is crucial for precision agriculture, improved yield estimation, and efficient resource management. This is especially impactful for smallholder farms, which characterize a vast portion of agricultural landscapes in the developing world and are essential to the livelihoods of billions. By enabling cost-effective and timely in-season monitoring, the technology can provide actionable insights to improve yields and mitigate losses, directly contributing to the economic well-being of these farmers and bolstering local and global food supplies.

\section{Additional Experimental Results}
\label{app:expts}

\subsection{Dataset Collection}
The India Dataset is a proprietary collection of images acquired over the period 2019-2022, covering varied agricultural topographies within India. 
We follow established precedent from works such as\cite{ai4}, \cite{sirko} in sourcing VHR imagery from Google Maps Satellite View. We take images sampled at 0.5 m GSD and from the following two satellite sources: Maxar Worldview, Airbus Pleiades.

Individual images represent a spatial extent of 320*320 $m^2$. Annotation was conducted by trained human personnel according to established protocols. An 'agricultural field' was operationally defined as a contiguous area meeting at least one of the following criteria: (1) cultivation of a single crop type, or (2) distinct visual boundaries separating it from neighboring parcels or non-agricultural features.
The Vietnam Dataset was constructed using publicly available labels provided by \citep{ai4}. As documented in \citep{ai4}, the high-resolution (HR) images at a given time t, denoted as $X_t$
were originally collected for August 2021. However, our preliminary analysis revealed inconsistencies between these provided labels and the corresponding $X_t$ imagery. Specifically, in some instances, reference images $X_r$ predating $X_t$ by six months or more exhibited greater concordance with the ground truth labels than the August 2021 images themselves. To ensure the integrity of our evaluation, the test set was subsequently filtered. This process retained only those image-label pairs where the August 2021 image ($X_t$) demonstrated clear and strong agreement with its associated ground truth labels, thereby excluding cases where labels appeared to reflect a significantly different temporal state. Table \ref{app:input_dets} summarises the properties of the datasets used.

\begin{table*}[h]
\centering
\caption{Summary of Input Data Characteristics}
\label{tab:input_data_summary}
\begin{tabular}{ p{2cm} p{8cm} p{8cm} } 
\toprule
\textbf{Category} & \textbf{Item} & \textbf{Detail / Specification} \\
\midrule
\multirow{3}{*}{\textbf{LR Inputs}} & Satellite Sources & Sentinel-2 L1C, Landsat 8/9, and Sentinel-1 GRD \\
& Spatial Resolution & $\mathbf{10\text{m}}$ GSD (Resampled) \\
& Temporal Span & $\mathbf{4}$ weeks before the target time ($t$) \\

& Embedding Dimension from $E_{LFM}$ & (128,128,64) $->$ (32,32,64) [After Cropping] \\
& Per Embedding Spatial Coverage & $320 \times 320 m^2$
\\

\midrule
\multirow{3}{*}{\textbf{HR Inputs}} & Spatial Resolution & $\mathbf{0.5\text{m}}$ GSD \\
& Embedding Dimension from $E_{HFM}$ & (120,120,3840) \\
& Per Embedding Spatial Coverage & $320 \times 320 m^2$
\\
& India Dataset Collection Time & 2019--2022 \\
& AI4SmallFarms Data Collection Time & August 2021 \citep{ai4} \\
\bottomrule
\label{app:input_dets}
\end{tabular}
\end{table*}

\subsection{Evaluation Metrics}
We use standard pixel-level metrics to evaluate the performance of our binary classification task.
Let $TP_i, TN_i, FP_i, FN_i$ denote the number of true positive, true negative, false positive, false negative pixels respectively in a single test image $i$.
For a test set with $N$ images the semantic IoU $IoU_S = \frac1N \sum_1^N \frac{TP_i}{TP_i+FP_i+FN_i}$.
At the instance level, we use the definition recommended for evaluating agricultural field instances (e.g., \citep{dua2024agriculturallandscapeunderstandingcountryscale}), viz.,
Instance IOU $IoU_I = \frac1M \sum_1^M (P_{m}^i \cap P^i)/(P_{m}^i \cup P^i)$, where $M$ is the number of ground truth instances in the test set,
$P_{m}^i$ is the total number of pixels in all the predicted $i^{\rm th}$ instances which have more than $t \%$ overlap with the  $i^{\rm th}$ ground truth instance and 
$P^i$ is the total number of pixels in the ground truth $i^{\rm th}$ instance. Accuracy, Precision, Recall and F1 score are computed from $TP_i, TN_i, FP_i, FN_i$ and averaged over the test set.

\subsection{Performance of Teacher and Human Labels}

The Teacher model predictions i.e. the predictions from the HR-FM on the HR images yield an $mIoU_S$ of 74.18 \% and an $mIoU_I$ of 71.17 \% for the India Dataset and $mIoU_S$ of 68.47 \% and an $mIoU_I$ of 63.29 \% for the Vietnam subset of the AI4SmallFarms Dataset.

\subsection{Ablations with Inference Timesteps}
\begin{figure*}
    \centering
    \includegraphics[width =0.75\linewidth]{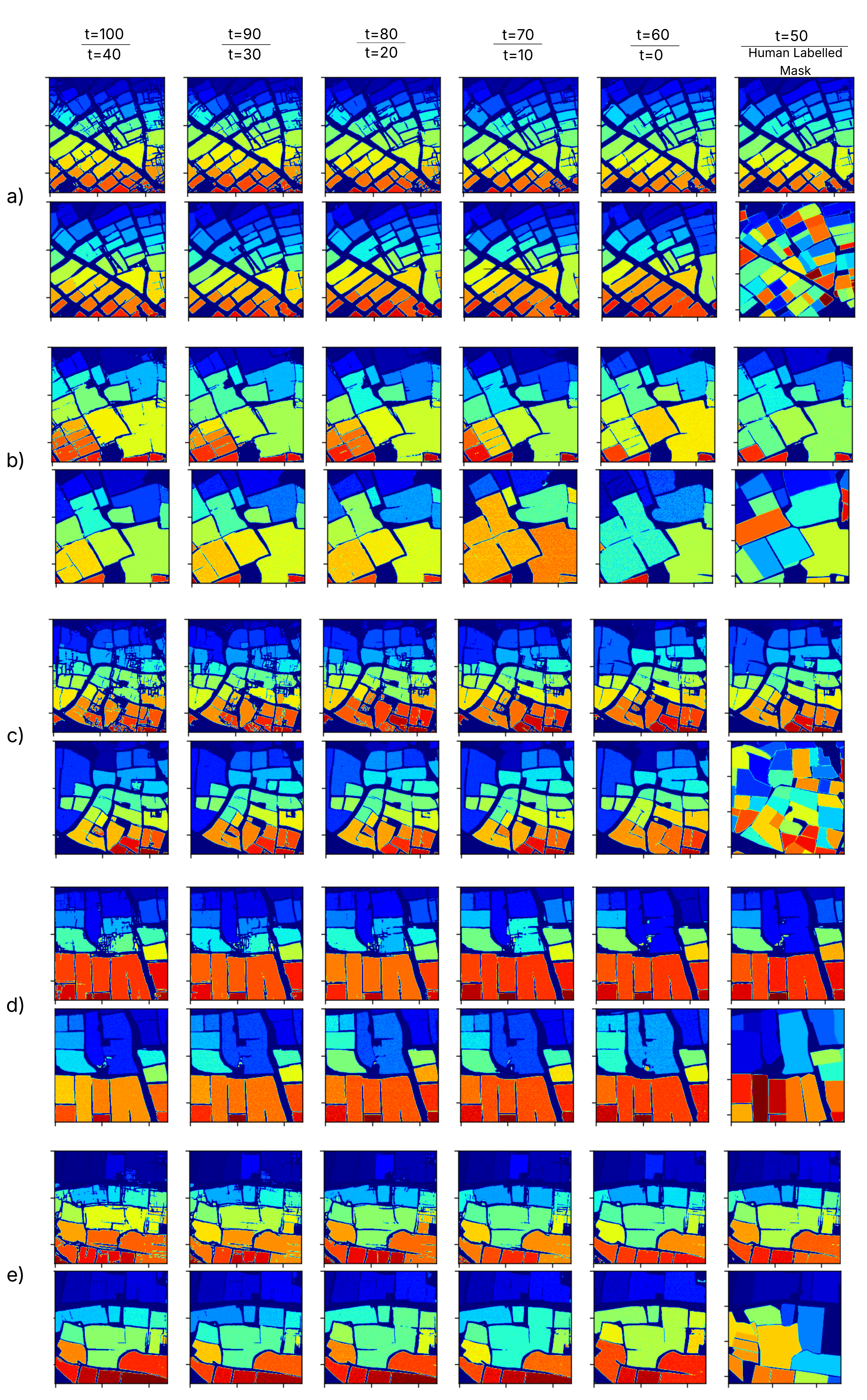}
    \caption{Qualitative results of last 100 denoising steps during inference for India Dataset.}
    \label{fig:t100}
\end{figure*}

Experimental results suggested that stopping the reverse diffusion process early produced better results.
Table \ref{tinf} displays the results for both India dataset and Vietnam dataset. Figure \ref{fig:t100} shows some qualitative examples from the last 100 denoising steps during inference time.

\begin{table*}[htbp] 
\captionsetup{font=footnotesize}
  \caption{Comparision of Performance on early stopping during Inference time for India and Vietnam Dataset. A: Accuracy, P: Precision, R: Recall}
\label{tab:segmentation-comparison-our-dataset} 
\centering
\scriptsize
\begin{tabular}{lcccccccccccc} 
\toprule
& \multicolumn{12}{c}{India Dataset} \\ 
\cmidrule(lr){2-13} 
& \multicolumn{6}{c}{Teacher Predictions} & \multicolumn{6}{c}{Human Labels} \\ 
\cmidrule(lr){2-7} \cmidrule(lr){8-13} 
Method & $mIoU_{I}$ & $mIoU_{S}$ & $A$ & $P$ & $R$ & $F1$ & $mIoU_{I}$ & $mIoU_{S}$ & $A$ & $P$ & $R$ & $F1$ \\ 
\midrule
$T_{Inf,500}$ & 49.16 & 72.44 & 84.89 & 60.49 & 61.27 & 58.67 & 48.21 & 70.71 &83.48 &59.80 & 57.28 & 56.58 \\
$T_{inf,470}$ & 58.44 & 74.96 & 86.33 & 62.11 &64.68 & 61.51 & 56.61 & 71.06 & 83.55 &60.94 & 59.38 & 58.34 \\ 
\bottomrule
\\
  \end{tabular}

  \begin{tabular}{lcccccccccccc} 
    & \multicolumn{12}{c}{Vietnam Dataset (AI4SmallFarms)} \\ 
\cmidrule(lr){2-7} \cmidrule(lr){8-13} 
    Method & $mIoU_{I}$ & $mIoU_{S}$ & $A$ & $P$ & $R$ & $F1$ & $mIoU_{I}$ & $mIoU_{S}$ & $A$ & $P$ & $R$ & $F1$ \\ 
    \midrule
    $T_{inf,500}$ & 49.14 & 72.30  & 82.10 & 76.68 & 77.92 & 75.92 & 48.77 & 68.20 & 79.11 & 72.77 & 78.30 & 72.59 \\
   $T_{inf,470}$  & 50.91  & 71.01 & 81.34 & 76.76 & 74.97 & 74.43 & 49.49 & 67.61 & 78.72 & 72.94 & 75.65 & 71.89   \\ 
    \bottomrule
      \\
  \end{tabular}

\label{tinf}
\end{table*}

\subsection{Qualitative results: Additional Examples}
\begin{figure*}
    \centering
    \includegraphics[width =0.85\textwidth]{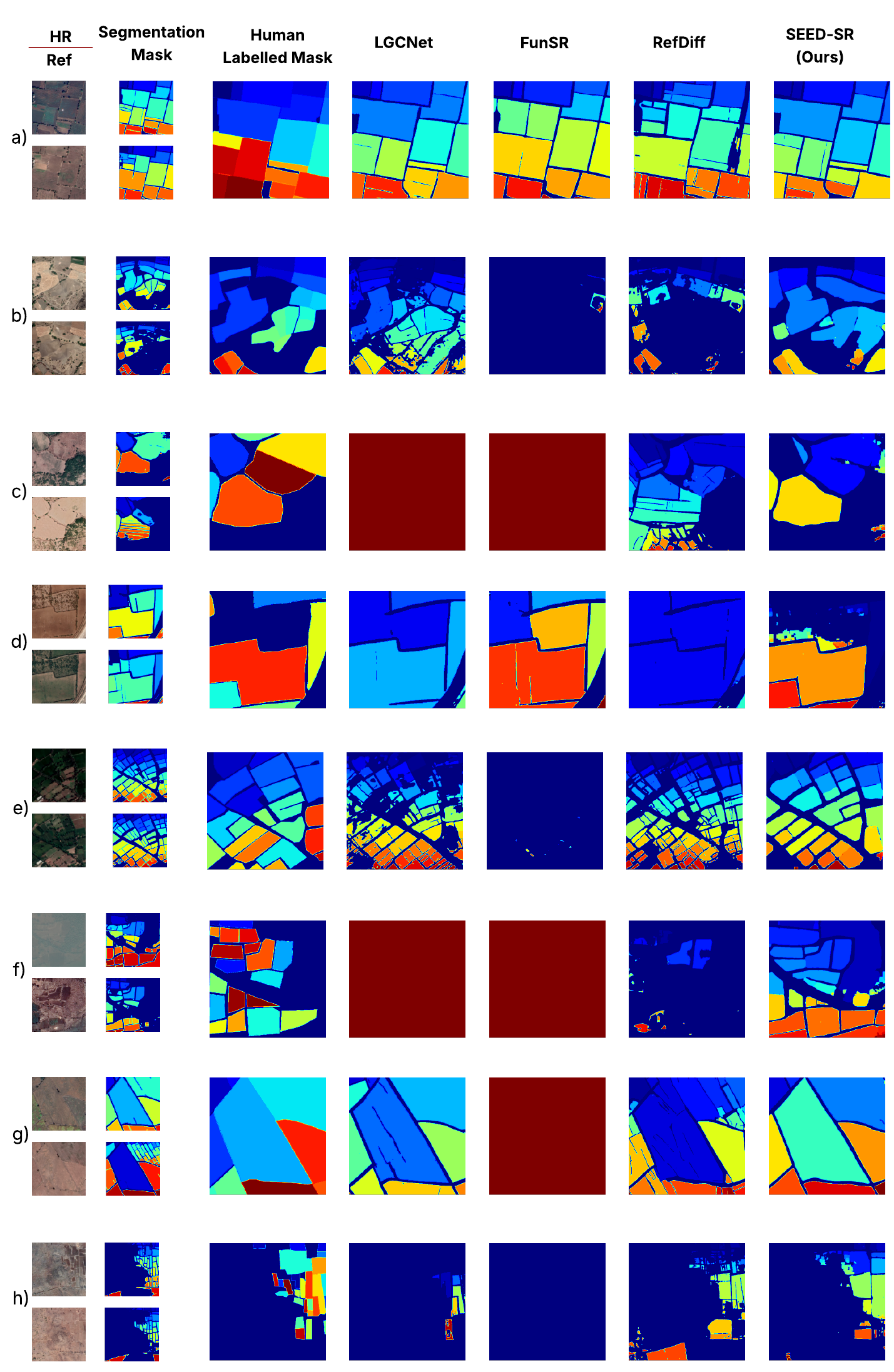}
    \caption{Additional Qualitative Examples from the India Dataset for Visual Comparison}
    \label{fig:addquals}
\end{figure*}

To provide a more nuanced understanding of our model's performance characteristics, Figure~\ref{fig:addquals} presents a selection of additional qualitative examples. These examples highlight the efficacy of our proposed method under certain conditions and its failure modes.

\begin{itemize}
    \item \textbf{Demonstrating Robustness to New Structures} Examples a,b showcase instances where our model successfully identifies and segments newly developed structures or significant emergent changes within the scene. This demonstrates our model's capability to detect objects or alterations not present in the reference image but from the LR inputs.

    \item \textbf{Superiority of Reference-Based Methods in Specific Contexts} In contrast, c,f,g illustrate specific scenarios where Ref-SR methods(RefDiff\citep{dong2024building} and SEED-SR(Ours) exhibit superior performance compared to non-reference-based approaches. These cases typically involve %
    subtle textural changes. 

    \item \textbf{Failure Mode:Logit Inconsistency} Example d reveals a specific failure mode observed in our model. In this instance, the introduction of noisy artifacts or erroneous detections in the output change map can be attributed to significant disagreement among the four distinct segmentation logits generated internally by our model prior to taking an average.

    \item \textbf{Comparative Advantage: Reduced Noise/Oversegmentation} Finally, e provides a comparative example where other existing methods exhibit common failure modes such as oversegmentation of changed regions or the introduction of spurious noisy artifacts in their predictions.
\end{itemize}

\begin{figure*}
    \centering
    \includegraphics[width =0.99\linewidth]{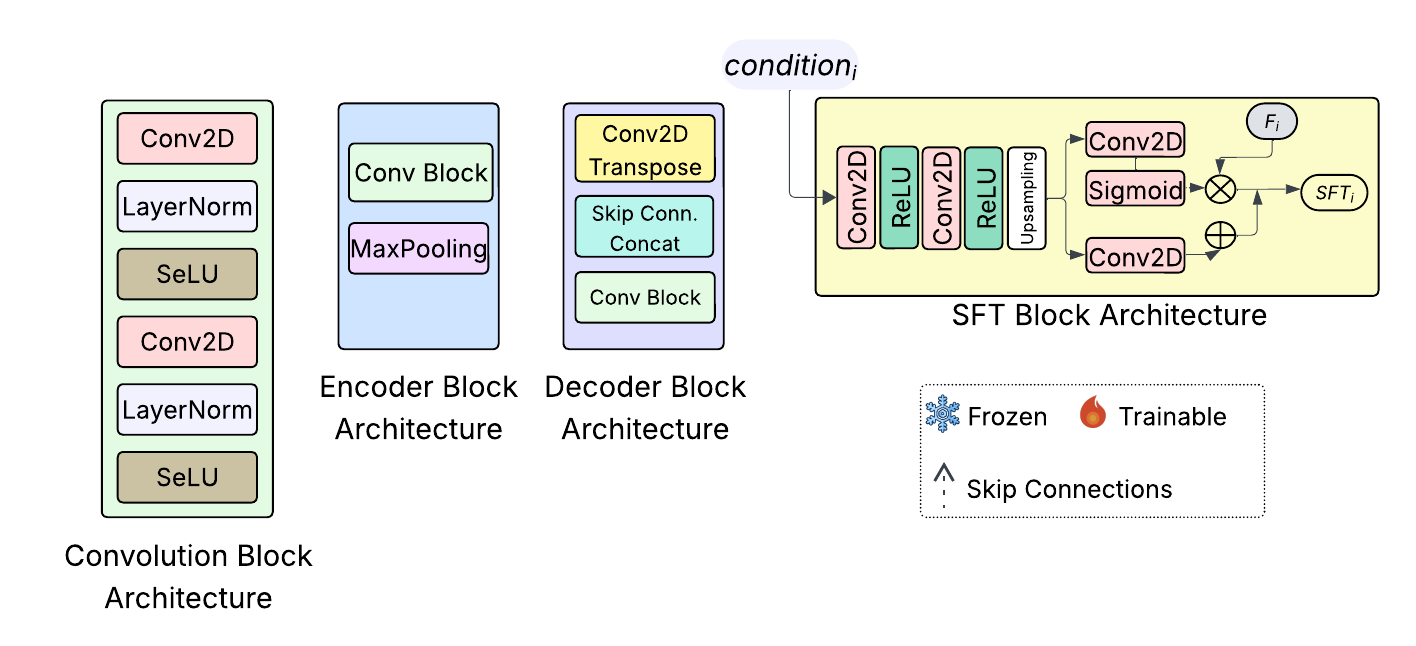}
    \caption{Architecture of ConvBlock, Encoder Block, Decoder Block and SFT as used in Figure~\ref{fig:model_training}}
    \label{fig:train_blocks}
\end{figure*}


\subsection{Evaluation of Segmentation Performance using SAM for Baseline Methods}
\label{app:sam}
\begin{figure*}
    \centering
    \includegraphics[width =1\textwidth]{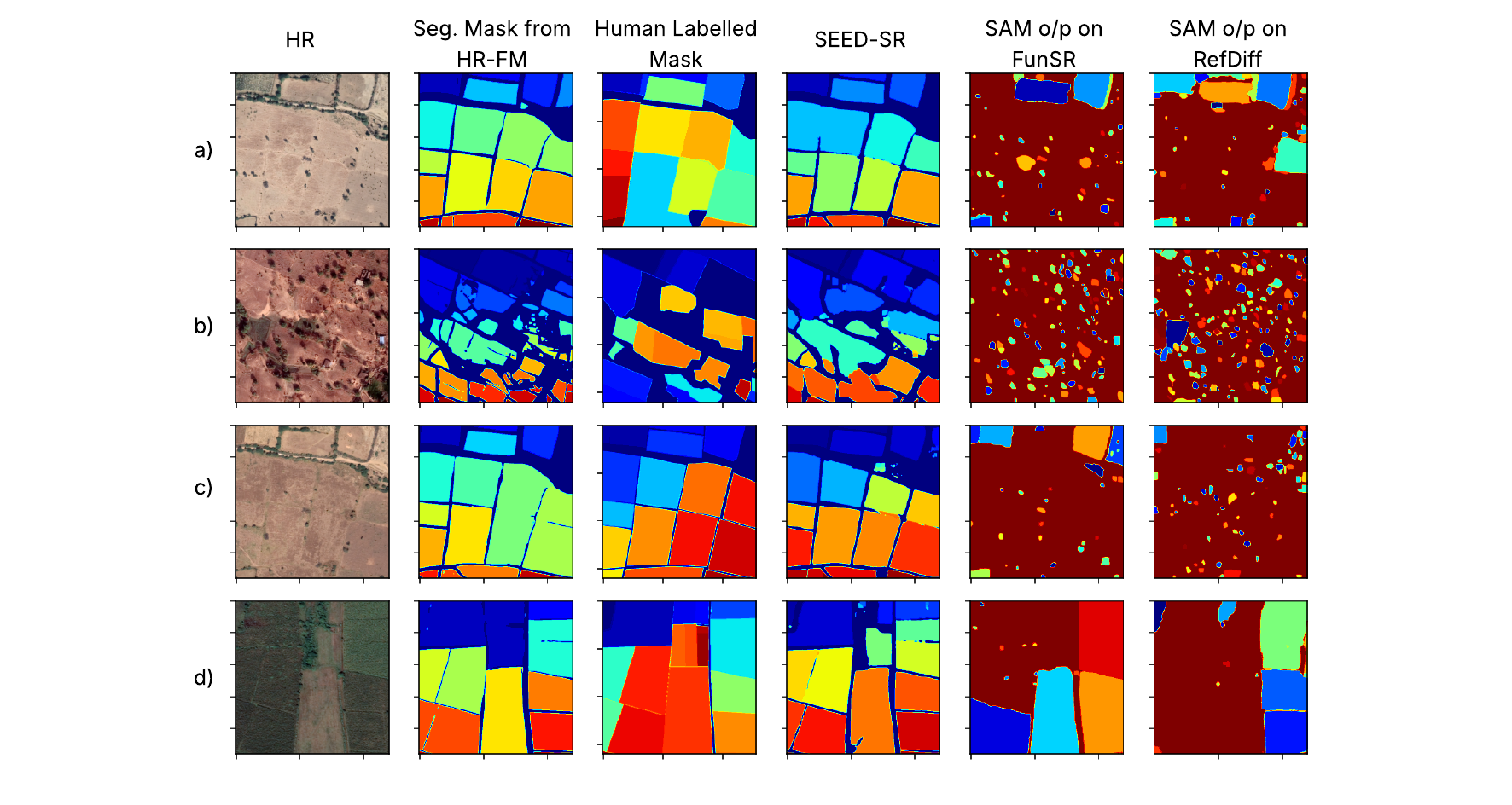}
    \caption{Zero Shot Instance Segmentation Outputs by SAM on Image Superresolution Baselines (RefDiff\citep{dong2024building} and FunSR\citep{funsr}) on the India Dataset}
    \label{fig:sam}
\end{figure*}

We present results in Figure \ref{fig:sam} for zero-shot performance of SAM\citep{kirillov2023segment} on the superresolved images generated by baseline methods and provide a qualitative comparison against segmentation results produced by HR-FM and SEED-SR's superresolved segmentation output. Visual results clearly demonstrate that the segmentation from HR-FM and SEED-SR are considerably more accurate and coherent for agricultural fields compared to zero-shot SAM, which largely struggles with delineating farm boundaries.

\subsection{Foundation Model Dependency and Failure Modes}
\label{app:failure}
SEED-SR's performance is coupled with the underlying FMs, a limitation we discuss in the conclusions. To further investigate this dependency, we isolated a subset of the test data where the per example mIoU of the teacher predictions (HR-FM’s performance on ground truth HR Images) against the human labels is poor (instance IoU $<$ 60\%).
This analysis confirms that SEED-SR's performance is bounded by the capabilities of the HR-FM. On the subset where the HR-FM performs poorly ($<$60\% IoU), HR-FM achieves 52.02 IoU, and SEED-SR achieves 37.56 IoU, as seen in Table \ref{tab:failure modes}. This expected behavior validates that the model successfully learns the mapping to the target latent space, but cannot correct for errors inherent in that space. 
Visual inspection shows that most of these images correspond to particularly difficult regions with few fields, or extremely small fields or dry/hilly terrains.
As expected, the best baseline method, RefDiff also performs poorly at $mIoU_I = 25.44$ in this set.

\begin{table}
    \centering
    \begin{tabular}{lc}\toprule
         \textbf{Method}& \textbf{$mIoU_{I}$ (vs. Human Labels)}
         \\\midrule
         HR-FM (Teacher prediction) & 52.02\\
    	SEED-SR Prediction & 37.56\\
         RefDiff Prediction & 25.44\\ \bottomrule
    \end{tabular}
    \caption{Performance on subset of the test data where the per example $mIoU_I$ of the teacher predictions( HR-FM’s performance on ground truth) against the human labels is poor ($mIoU_I$ $<$ 60\%)}
    \label{tab:failure modes}
\end{table}

\subsection{Computational Cost}
\label{app:time}

Our model runs for 200,000 training steps. Each training step for a batch size of 128 takes approximately 1 second over 8 Google TPUv5(s). The complete training framework took 58.3 hours. The complete inference framework along with metrics computation took approximately 3 hours. The inference step takes 2.64 seconds per denoising step for a batch size of 320.

Table \ref{tab:inference_time} shows the average inference time per example at maximum batch size for SEED-SR and three baselines.
While diffusion methods (SEED-SR, RefDiff) are slower than non-generative methods (LGCNet, FunSR), the substantial performance gain justifies this trade-off for high-accuracy applications. Notably, SEED-SR (3.968s) is slightly faster than the comparable SOTA diffusion baseline, RefDiff (4.12s). The average inference time per example at maximum batch size was calculated at maximum throughput to take advantage of the hardware's parallelization capabilities across models.

\begin{table}[h]
\centering
\caption{Model Inference Time Comparison}
\label{tab:inference_time}
\begin{tabular}{lcc}
\toprule
\textbf{Method} & \textbf{Average Inference time per Example} \\
\midrule
SEED-SR(Ours) & 3.968 s  \\
RefDiff \citep{dong2024building} & 4.12 s  \\
FunSR \cite{funsr} & 0.7 s  \\
LGCNet \cite{7937881} & 0.64 s & \\
\bottomrule
\end{tabular}
\end{table}




\subsection{Choice of different number of masks for the production of the final segmentation map}
\label{app:masks}
Diffusion models are heavily reliant on the random seed for the variability in examples and often a mean of predictions is taken to reduce stochastic variance. The masks were computed for the validation set of the India dataset(291 images). Table \ref{tab:mean_masks} shows the performance for different number of masks. The masks were randomly chosen for each example. The results show clear gains up to N=4, with diminishing returns thereafter. We selected N=4 as the optimal balance between performance ($\approx 2.3$ $mIoU_I$ gain over N=2) and inference cost.

\begin{table}[htbp]
\centering
\footnotesize
\caption{Sensitivity to number of masks}
\label{tab:mask_ablation}
\begin{tabular}{{ p{2.5cm}  p{2.4cm}  p{1.8cm} }}
\toprule
\textbf{No. of Masks} & \textbf{Teacher Predictions ($mIoU_{I}$)} & \textbf{Human Labels ($mIoU_{I}$)} \\
\midrule
1 & 49.9693 & 49.7525 \\
2 & 55.2753 & 53.8581 \\
3 & 55.5124 & 54.6811 \\
\textbf{4} & \textbf{57.7448} & \textbf{56.0503} \\
5 & 57.7419 & 56.0247 \\
6 & 57.6382 & 56.4666 \\
7 & 57.8608 & 56.5775 \\
8 & 58.3667 & 56.5203 \\
\bottomrule
\label{tab:mean_masks}
\end{tabular}
\end{table}

\section{Additional Model Details}
\label{app:model_details}

\subsection{Novelty in DDPM Architecture}
\label{app:novel}
Our contribution is the novel integration of two distinct geospatial FMs via a latent diffusion model for 20× task-specific SR. This is non-trivial due to the high-dimensional embeddings (120×120×3840) . The novelty in our DDPM architecture \cref{dims} includes the integration strategy for multi-modal conditioning (LR and Ref) using self-attention on concatenated inputs and introducing a convolution block to downsample the feature map spatially (to 7x7) before applying cross-attention at the bottleneck (L207-217); at extremely high dimensions. This innovation reduced memory usage from ~14 GB to ~2.75 MB and FLOPs from 423 GFLOPs to 5 MFLOPs (See Table), making diffusion feasible in this feature space.

\begin{table}[htbp]
\centering
\footnotesize
\caption{Computational Cost Comparison justifying the design choice of DDPM architecture}
\label{tab:computational_cost}
\begin{tabular}{lcc}
\toprule
\textbf{Metric} & \textbf{Case A (7$\times$7)} & \textbf{Case B (120$\times$120)} \\
\midrule
Tokens per sample & 49 & 14,400 \\
Attention memory & $\sim$0.15 MB & $\sim$13.2 $\times$ 10$^3$ MB (float32) \\
Total memory & $\sim$2.75 MB & $\sim$14 $\times$ 10$^3$ MB \\
FLOPs & $\sim$5 MFLOPs & $\sim$423 $\times$ 10$^3$ MFLOPs \\
\bottomrule
\end{tabular}
\end{table}
 
\subsection{Experimental Details}

SEED-SR was trained using the AdamW optimizer \citep{adamw} with a learning rate $5 \times 10^{-4}$, a batch size of 128, for a total of $200,000$ steps. 
A total of $T=500$ timesteps were employed for the forward process and $T=470$ steps were employed for the reverse diffusion processes.
The noise variance schedule for the diffusion process was linearly interpolated from $\beta_1 = 10^{-4}$ to $\beta_T = 0.02$. 
The training was done on Google's TPUv5. 


\subsection{Diffusion Process}

\subsubsection{Forward Processs}

The forward process gradually adds Gaussian noise to the HR embedding $e_h$ over T discrete timesteps. Given $e_{h,0}$ the noised embedding $e_{h,\tau}$ at timestep $\tau \epsilon [0,T]$ is obtained by sampling from the distribution:

$$q(e_{h,\tau} | e_{h,0}) = \mathcal{N}(e_{h,\tau}; \sqrt{\bar{\alpha}_{h,0}, (1 - \bar{\alpha}_{\tau})\mathbf{I}})$$

where  $$e_{h,\tau} = \sqrt{\bar{\alpha}_{\tau}} e_{h,0} + \sqrt{1 - \bar{\alpha}_{\tau}} \varepsilon, for  \varepsilon  \approx N(0,I)$$

The variables $\alpha_\tau$ =$ 1 - \beta_\tau$ and  $\bar{\alpha}_{\tau}$ = ${\prod_{i=1}^{\tau}} \alpha_i$, where ${{\beta_\tau}^T}_{\tau=1}$ is a predefined noise variance schedule.

\subsubsection{Reverse Denoising Process}

The reverse process learns to recover the original clean embedding $e _{h,0}$ from its noised version $e_{h,\tau}$ conditioned on $e _l, e_r$ and the timestep $\tau$. This is achieved by training a neural network $G_{\theta}(e_{h,\tau}, \tau, c)$ to predict an estimate of the clean embedding,  $\hat{e}_{h,0}$ = $G_{\theta}(e_{h,\tau}, \tau, c)$, where c=condition($e_l,e_r$) represents the combined conditioning information. The model is trained by minimizing the mean squared error between the true clean embedding and its prediction:

$$L_{\text{diffusion}} = \mathbb{E}_{\tau, e_{h,0}, \varepsilon} \left[ || e_{h,0} - G_{\theta}(e_{h,\tau}, \tau, c) ||^2_2 \right]$$

The sampling process proceeds as below:

We begin by sampling $e_{h, \tau}$ from a standard isotropic Gaussian distribution,

$$e_{h, \tau} \sim \mathcal{N}(0,\mathcal{I})$$

For each timestep $\tau$ from T down to 1, to obtain $e_{h, \tau-1}$ from $e_{h, \tau}$ we estimate $\hat{e}_{h, 0}$ from the model $G_{\theta}$,

$$\hat{e}_{h, 0} = G_{\theta}(e_{h,\tau}, \tau, c) $$

Using the predicted $\hat{e}_{h, 0}$ and the current $e_{h,\tau}$, the mean of the posterior distribution $q(e_{h, \tau-1} | e_{h, \tau}, \hat{e}_{h,0})$ denoted by $\tilde{\mu}_t(e_{h, \tau}, \hat{e}_{h,0})$ is calculated as:

$$\tilde{\mu}_t(e_{h, \tau}, e_{h,0}) =  \frac{\sqrt{\bar{\alpha}_{\tau-1}}\beta_\tau}{1-{\bar{\alpha}_\tau}}\hat{e}_{h,0}
+ \frac{\sqrt{\alpha_{\tau}}(1-\bar{\alpha}_{\tau-1}}{1-{\bar{\alpha}_\tau}}e_{h, \tau}
$$

The sample for the previous timestep is then drawn by adding scaled Gaussian noise to the posterior mean:

$$e_{h, \tau-1} = \tilde{\mu}_t(e_{h, \tau}, e_{h,0}) + \sigma_tz$$

where $z \sim \mathcal{N}(0,\mathcal{I})$ if $\tau > 1$ and z=0 if $\tau=1$. The choice of noise variance $\sigma_{\tau}^2 = \beta_\tau$ follows from \citep{ddpm}. So the sampling step becomes, 

$$e_{h, \tau-1} = \tilde{\mu}_t(e_{h, \tau}, e_{h,0}) + \sqrt{\beta_\tau}z$$

Thus the $\hat{e}_{h,0}^{final}$ is the generated high-resolution embedding, which is then passed to the frozen decoder $D_{HSM}$ to produce the final segmentation map.





\subsubsection{Spatially Feature Transform (SFT) blocks}

To specifically introduce and leverage the distinct characteristics of the low-resolution input ($e_l$) and the reference input ($e_r$), we employ Spatially Feature Transform (SFT) blocks at different stages within the decoder. Figure \ref{fig:train_blocks} shows a detailed architecture of the SFT Block, Convolution Block, Encoder Block and Decoder Block as shown in Figure \ref{fig:model_training}.  Architecturally, each SFT block first processes its conditioning input through a pair of 3x3 convolutional layers with ReLU activations (transforming channels to 128, then 64).

Let $c_{in}$ be the input conditioning in the SFT block, the convolutional block followed by ReLU produces,
$$C_1 = \text{ReLU}(\text{Conv}_{128, 3 \times 3}(c_{in}))$$

Progressively we apply another convolutional block to reduce the channels to 64,
$$C_2 = \text{ReLU}(\text{Conv}_{64, 3 \times 3}(C_1))$$

The output $C_2$ is then bilinearly upsampled to match the spatial dimensions (H,W) of the main feature map F that it will modulate, resulting in $C_{up}$. Subsequently, from the upsampled conditioning features $C_{up}$, two separate 3x3 convolutional branches, are used to predict spatial-specific modulation parameters: a scale factor ($\gamma$) and a shift factor ($\beta$). Both $\gamma$ and $\beta$ will have the same number of channels ($n_F$) as the main feature map F. 

$$\gamma = \sigma(\text{Conv}_{n_F, 3 \times 3}(C_{up}))$$

$$\beta = \text{Conv}_{C_F, 3 \times 3}(C_{up})$$

Finally, the learned scale ($\gamma$) and shift ($\beta$) parameters are applied to the main feature map F using an element-wise affine transformation. The feature map F is first scaled by $\gamma$ and then the shift $\beta$ is added such that,
$$F_{out} = \gamma \odot F + \beta$$

\subsection{Layerwise Dimensions in Our Model}

Table 5 displays the layerwise dimensions of SEED-SR's denoising diffusion model.

\begin{table}[htbp] 
\caption{Layerwise Dimensions of our Model}
  \centering
    \footnotesize
  \begin{tabular}{lc} 
    \toprule
    Layer & Dimension \\ 
    \midrule
    Input Noised HR & (120,120,3840)  \\
    Input Ref & (120,120,3840) \\
    Input LR & (32,32,64)  \\
    Input Timestep & (1,) \\
    Usampled LR & (120,120,64)  \\
    Concatenated Encoder Input & (120,120,7744)\\
    Encoder Block 1 & (60,60,320)\\
    Encoder Block 2 & (30,30,480)\\
    Encoder Block 3 & (15,15,560)\\
    Encoder Block 3 with Timestep Embedding & (15,15,560)\\
    Encoder Block 4 & (7,7,640)\\
    Attention Block & (7,7,640)\\
    Bottleneck Conv Block & (7,7,784)\\
    Decoder Block 1 & (15,15,640)\\
    SFT Block 1 & (15,15,640)\\
    Decoder Block 2 & (30,30,560)\\
    Decoder Block 3 & (60,60,480)\\
    SFT Block 2 & (60,60,480)\\
    Decoder Block 3 with SFT and Timestep Embedding & (60,60,480)\\
    Decoder Block 4 & (120,120,320)\\
    Attention Block & (120,120,320)\\
    Convolution Block & (120,120,3840)\\

    \bottomrule
  \end{tabular}
  
  \label{layer}
\end{table}

\subsection{Changes in Ablation studies}

Table 6 specifies the layerwise changes in model dimensions because of the different inputs of ablation studies.
\begin{table}[htbp] 
    \centering 
    \caption{Layerwise Input Dimension Changes for Ablation Studies}
    
    \begin{subtable}[t]{0.48\textwidth} 
        \centering
        \caption{Config A: S2 + Ref, no embeddings}
        \label{tab:sub_table_a}
         \begin{tabular}{lc} 
    \toprule
    Layer & Dimension \\ 
    \midrule
    Input Noised HR & (480,480,3)  \\
    Input Ref & (480,480,3) \\
    Input LR & (32,80,80,12)  \\
    Input Timestep & (1,) \\
    Reshaped and Upsampled LR & (480,480,384)  \\
    Concatenated Encoder Input & (480,480,390)\\
    \makecell[c]{$\vdots$} & \makecell[c]{$\vdots$} \\

    Convolution Block & (480,480,3)\\

    \bottomrule
  \end{tabular}
    \end{subtable}
    \hfill 
    \begin{subtable}[t]{0.48\textwidth} 
        \centering
        \caption{Config B: Only S2 images, no embeddings}
        \label{tab:sub_table_b}
          \begin{tabular}{lc} 
    \toprule
    Layer & Dimension \\ 
    \midrule
    Input Noised HR & (120,120,3840)  \\
    Input LR & (32,80,80,12)  \\
    Input Ref & NA \\
    Input Timestep & (1,) \\
    Upsampled LR & (120,120,384)  \\
    Concatenated Encoder Input & (120,120,4224)\\
    \makecell[c]{$\vdots$} & \makecell[c]{$\vdots$} \\

    Convolution Block & (120,120,3840)\\

    \bottomrule
  \end{tabular}
    \end{subtable}
    \end{table}
\begin{table}[htbp]
    \begin{subtable}[t]{0.48\textwidth} 
        \centering
        \caption{Config C: Only LR Embeddings}
        \label{tab:sub_table_b}
          \begin{tabular}{lc} 
    \toprule
    Layer & Dimension \\ 
    \midrule
    Input Noised HR & (120,120,3840)  \\
    Input LR & (32,32,64)  \\
    Input Ref & NA \\
    Input Timestep & (1,) \\
    Upsampled LR & (120,120,64)  \\
    Concatenated Encoder Input & (120,120,3904)\\
    \makecell[c]{$\vdots$} & \makecell[c]{$\vdots$} \\

    Convolution Block & (120,120,3840)\\

    \bottomrule
  \end{tabular}
    \end{subtable}
    \hfill
\begin{subtable}[t]{0.48\textwidth} 
        \centering
        \caption{Config D: Only Ref Embeddings}
        \label{tab:sub_table_b}
   \begin{tabular}{lc} 
    \toprule
    Layer & Dimension \\ 
    \midrule
    Input Noised HR & (120,120,3840)  \\
    Input LR & NA \\
    Input Ref & (120,120,3840) \\
    Input Timestep & (1,) \\
    Upsampled LR & NA \\
    Concatenated Encoder Input & (120,120,7680)\\
    \makecell[c]{$\vdots$} & \makecell[c]{$\vdots$} \\
    Convolution Block & (120,120,3840)\\

    \bottomrule
  \end{tabular}
    \end{subtable}
    \end{table}
    \begin{table}[h]
    \centering
    \begin{subtable}[t]{0.48\textwidth} 
    
        \caption{Config E: Ref Embeddings with S2 Images}
        \label{tab:sub_table_b}
    \begin{tabular}{lc} 
    \toprule
    Layer & Dimension \\ 
    \midrule
    Input Noised HR & (120,120,3840)  \\
    Input Ref & (120,120,3840) \\
    Input LR & (32,80,80,12) \\
    Input Timestep & (1,) \\
    Reshaped and Upsampled LR & (120,120,384) \\
    Concatenated Encoder Input & (120,120,8064)\\
    \makecell[c]{$\vdots$} & \makecell[c]{$\vdots$} \\

    Convolution Block & (120,120,3840)\\

    \bottomrule
  \end{tabular}
    \end{subtable}
\label{tab:main_table_label}
\end{table}

\end{document}